\documentclass{article}

 \usepackage[main, final]{neurips_2026}

\usepackage[utf8]{inputenc} 
\usepackage[T1]{fontenc}    
\usepackage[hypertexnames=false]{hyperref}       
\usepackage{url}            
\usepackage{booktabs}       
\usepackage{amsfonts}       
\usepackage{nicefrac}       
\usepackage{microtype}      
\usepackage{xcolor}         

\usepackage{graphicx}
\usepackage{xspace}
\usepackage{enumitem}
\usepackage{amsmath}
\usepackage{amssymb}
\usepackage{bm}
\usepackage{booktabs}
\usepackage{tabularx}
\usepackage{float}
\usepackage{pifont}
\usepackage{siunitx} 
\usepackage[colorinlistoftodos]{todonotes}
\usepackage{stfloats}
\newcommand*\iftodonotes{\if@todonotes@disabled\expandafter\@secondoftwo\else\expandafter\@firstoftwo\fi}
\makeatother

\newcommand{\eclektic}{ECLeKTic\xspace}
\newcommand{\multiloko}{MultiLoKo\xspace}
\newcommand{\gmmlu}{Global MMLU\xspace}
\newcommand{\glm}{GLM-4.5-Air\xspace}
\newcommand{\qwen}{Qwen3-30B-A3B\xspace}

\newcommand{\gemini}{Gemini\xspace}
\newcommand{\methodname}{XICI\xspace}
\newcounter{rownum}
\newcommand{\rownumber}{\stepcounter{rownum}\small{\arabic{rownum}}}

\title{Knowledge Localization in Mixture-of-Experts LLMs \\ Using Cross-Lingual Inconsistency}

\author{
  \begin{tabular}{ccc}
    Lucas Bandarkar\textsuperscript{\ding{70}}$^\dagger$ & Alan Ansell\textsuperscript{\ding{70}} & Trevor Cohn\textsuperscript{\ding{70}}$^\phi$ \\
    \multicolumn{3}{c}{\normalfont \textsuperscript{\ding{70}}Google Research} \\
    \multicolumn{2}{l}{$^\dagger$\normalfont University of California, Los Angeles} & $^\phi$\normalfont University of Melbourne 
  \end{tabular}%
}

\begin{document}

\maketitle

\begin{abstract}
Modern LLMs continue to exhibit significant variance in behavior across languages, such as being able to recall factual information in some languages but not others. While typically studied as a problem to be mitigated, in this work, we propose leveraging this cross-lingual inconsistency as a tool for interpretability in mixture-of-experts (MoE) LLMs.
Our knowledge localization framework contrasts routing for sets of languages where the model correctly recalls information with languages where it fails. This allows us to isolate model components that play a functional role in recalling a piece of knowledge.
Our method proceeds in two stages: (1) querying the model with difficult factual questions across a diverse set of languages to generate ``success'' and ``failure'' activation buckets and then (2) applying a statistical contrastive analysis to the MoE router logits to identify knowledge-centric experts. To verify that this small group of experts is needed to answer a knowledge-seeking question, we deactivate them 
and re-ask the question. We find that despite only deactivating about 20 out of 6000 experts, the model no longer answers correctly in over 40\% of cases.
Generally, this method provides a realistic and scalable localization approach to increasingly complex LLMs, but also suggests redundant and highly dispersed knowledge parameterization in MoEs.
\end{abstract}

\begin{figure}[H] 
    \centering
    \includegraphics[width=\columnwidth]{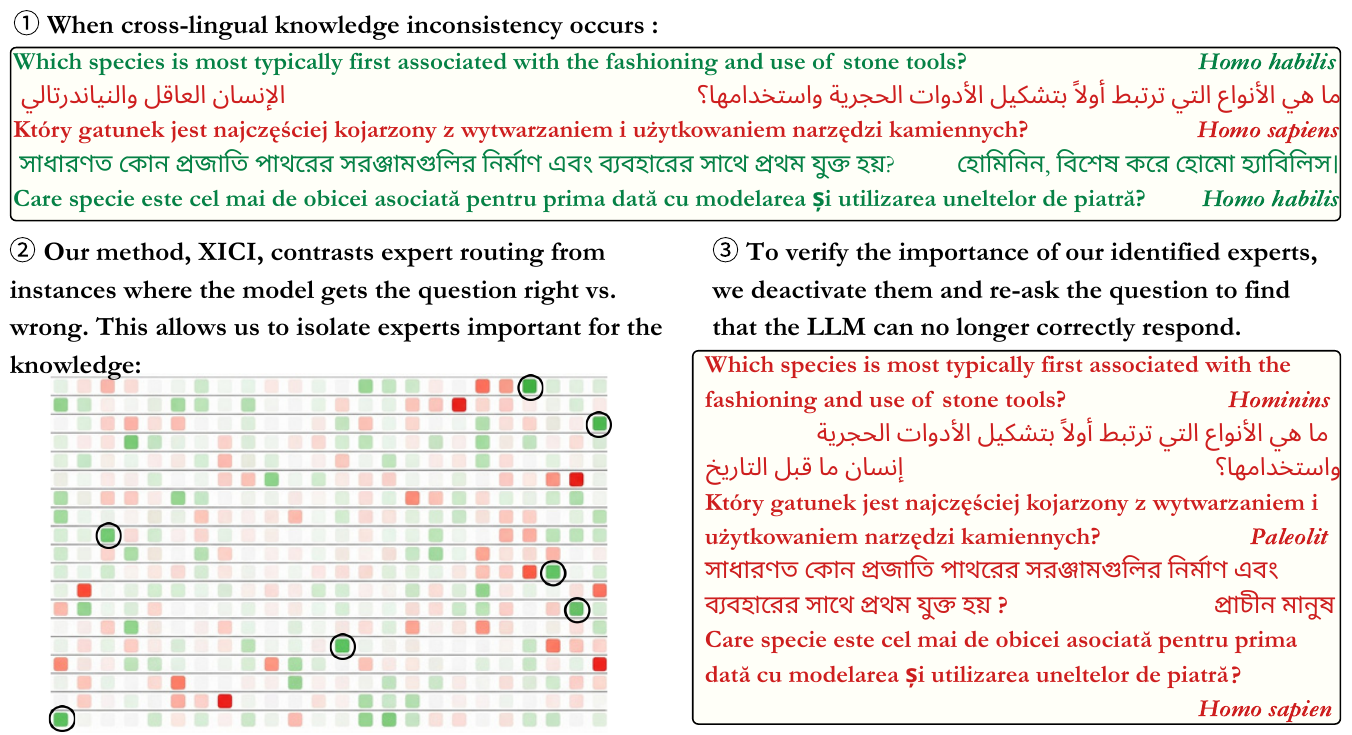}
    \caption{Overview of our knowledge localization framework for MoE LLMs that uses Cross-Lingual Inconsistency for Contrastive Identification (\textbf{XICI}).}
    \label{fig:top_right_image}
\end{figure}

\section{Introduction}

The accelerated development of large language models (LLMs) has far outpaced our ability to understand them. Despite significant effort, modern interpretability research has struggled to provide a verifiable understanding of how LLMs store and retrieve knowledge, something they do incredibly well. The field of mechanistic interpretability, for example, has focused on reverse-engineering models at neuron-level, but such ``bottom-up'', ultra-granular approaches have failed to reliably explain the increasing complexity of modern LLMs \citep{hendrycks2024misguided, nanda2025pragmatic}. A more tractable starting point may be knowledge localization — the task of identifying which parts of a model are responsible for retrieving a given fact \citep{hase2023does}. While often motivated by applications such as model editing or unlearning, knowledge localization is also a foundational scientific problem for studying how knowledge is learned and parameterized. Yet current localization methods are unreliable and difficult to validate, which calls for a search for very different strategies.

The recent transition of frontier LLMs from dense models to sparse mixture-of-experts (MoE) architectures \citep{shazeer2017} has only added to the challenges. In MoEs, the MLP block of each decoder layer is replaced with $E$ small blocks (\textit{experts}) where only $k \ll E$ are activated for each token. This sparsity enables much larger and better models, but relies on non-differentiable, discrete activations and results in highly distributed and redundant representations. As a result, many interpretability methods have become even less feasible. However, sparse activations also provide a natural unit of abstraction: the experts. This modularity offers a scalable lens for interpretability, uncovering structured patterns such as expert specialization for various functionality, domains, and languages \citep{muennighoff2025olmoe, bandarkar2026multilingual, li2026understanding, fayyaz2026steering}.

In this work, we propose a new angle for dissecting models rooted in the common observation that LLMs often behave differently when prompted with semantically equivalent inputs across languages. Modern LLMs operate in abstract representation spaces that are accessible via many input languages, allowing them to transfer knowledge and reasoning capabilities. However, the language-universality of these parameters is far from perfect and accessing this knowledge with a high rate of dependability remains difficult. As a result, we propose leveraging unsuccessful responses as an instrument for interpretability, rather than requiring unnatural perturbations. That is, we prompt semantically equivalent inputs in many languages to an LLM and if it produces the desired output in some languages but not others, we have two sets of computation graphs that we can contrast for analysis — a framework we call 
Cross-lingual Inconsistency for Contrastive Identification (XICI).


We apply XICI to MoEs for expert-level knowledge localization. With hundreds of factual-recall prompts translated across numerous languages from \eclektic \citep{eclektic}, \multiloko \citep{hupkes2025multiloko}, and the prehistory subject from \gmmlu \citep{singh-etal-2025-global}, we collect positive and negative router data from the MoE block at every layer. 
We then use the Mann-Whitney U-test to statistically separate activations \textit{common} amongst languages where the model answered correctly and \textit{missing} from incorrect forward passes. On top of this statistical test, we take numerous additional steps to ensure experts are important for question-specific knowledge, rather than confounding factors. The result of this attribution method is a small set of experts that we identify as being important for a factoid.


To verify the importance of such experts, we deactivate them (causal ablation) without disrupting the natural distribution of hidden states and evaluate the LLM's ability to answer such questions without the necessary experts. Across two models and over 400 questions, we find that when deactivating about 0.3\% of experts across the model, we achieve an ablation success rate of over 40\%.
Given the distributed nature of state-of-the-art MoEs, the ablation success rate and its comparison to baselines suggest significant promise in our approach.

We start by discussing the background and intuition behind this methodology in Sections~\ref{relatedwork} and~\ref{inconsistency}. In Section~\ref{identification}, we detail our XICI implementation. In Section~\ref{intervention}, we present the outcome of causally ablating the identified experts for each question. Finally in Section~\ref{discussion}, we discuss takeaways about XICI and broadly, knowledge in MoEs.
\section{Related Work} \label{relatedwork}

\subsection{Knowledge Localization and Editing}

Significant recent interpretability research in LLM knowledge has focused on mechanistic analyses of the model weights.
While the locate-then-edit paradigm \citep{meng2022locating} initially promised a surgical understanding of LLM parameters, recent efforts reveal significant gaps in current knowledge localization \citep{sharkey2025open}. Methods like Integrated Gradients \citep{sundararajan17a, lundstrom22a} and Causal Tracing \citep{meng2022locating, hase2023does} often suffer from low precision. Even as Sparse Autoencoders (SAEs) \citep{huben2024sparse} attempt to disentangle polysemanticity into interpretable features, they still fail to reveal verifiable mechanisms that cause model behavior.
As a result of these challenges, numerous researchers have advocated against continued attempts to reverse-engineer LLMs \citep{hendrycks2024misguided, sharkey2025open, nanda2025pragmatic}.
Regardless, knowledge has been consistently linked to FFN layers \citep{meng2022locating, pmet, geva-etal-2023-dissecting}, likely due to their large parameter capacity. These complex networks, or knowledge circuits, are typically distributed across many layers \citep{yao2024knowledge}.

\subsection{Interpretability in Mixture-of-Experts}

As discussed in the introduction, many localization methods designed for dense models are not easily adaptable to MoEs \citep{gu2026moeedit, li-etal-2025-decoding} (e.g. circuit analysis). In particular, \citet{wang2026deconstructingpretrainingknowledgeattribution} claims knowledge localization is more challenging in MoEs as sparsity fosters even more distributed knowledge storage. However, there is growing evidence that MoEs learn to specialize experts by function or semantics, providing interpretability research with convenient modular units \citep{muennighoff2025olmoe, lo-etal-2025-closer, olson-etal-2025-probing,bandarkar2026multilingual}. Evidence suggests this specialization is established quite early in pretraining \citep{xue2024openmoeearlyeffortopen, wang2026deconstructingpretrainingknowledgeattribution} and is more pronounced with global-batch load-balancing \citep{qiu-etal-2025-demons}. Functionality arises from numerous experts collaborating across layers \citep{ying2025benchmarksunderstandingmixtureofexpertsmodels}, but \citet{fayyaz2026steering} demonstrates that individual experts can be identified and deactivated to reliably steer model behavior.

\subsection{Language-Agnostic Representations}
Fundamentally, our method leverages LLMs' language-agnostic representations. These representations, attainable from input in any language, have been studied extensively and found to mostly reside in middle model layers \citep{kojima-etal-2024-multilingual,tang-etal-2024-language,wu2025the,dumas-etal-2025-separating}. As LLMs have scaled, this functional dissociation between linguistic capabilities and other capabilities has become stronger and more modular \citep{qiu-etal-2024-unlocking,chen2025the,bandarkar2025unreasonable}, replicating patterns in the human brain \citep{MAHOWALD2024517}. Correspondingly, \citet{wang-etal-2025-lost-multilinguality} and \citet{sun_aaai} find clear evidence that knowledge parameterization is language-independent. That being said, \citet{liu-etal-2025-tracing} observe widespread exceptions to this, especially with less common knowledge. Regardless, cross-lingual knowledge editing remains notoriously challenging \citep{nie-etal-2025-bmike}. Recent works find that inconsistent outputs across languages often arises from errors in the implicit cross-lingual mapping that occurs in the first and last model layers \citep{lu-etal-2025-paths,wang-etal-2025-lost-multilinguality, sun_aaai}, including in MoEs \citep{bandarkar2026multilingual}. The method defined in this work relies on these findings that (1) knowledge encoding is mostly language-agnostic and (2) inconsistency arises from anisotropy that causes routing variation across languages.
\section{Setup} \label{setup}

In this work, we study two state-of-the-art deep and sparse MoE LLMs: \glm \citep{glm} (which is 106B parameters with 12B active) and \qwen \citep{qwen3} (specifically, \href{https://huggingface.co/Qwen/Qwen3-30B-A3B-Instruct-2507}{\qwen-Instruct-2507}).

We source knowledge-centric, factual-recall-style questions with pre-existing translations from the multilingual datasets \eclektic \citep{eclektic}, \multiloko \citep{hupkes2025multiloko}, and the prehistory subject from \gmmlu \citep{singh-etal-2025-global}. \eclektic consists of 383 difficult  local knowledge questions sourced from 12 languages and machine-translated (with verification)
to be 12-way parallel. \multiloko is sourced similarly, but we specifically only use the 250 questions in the English-source split, as it is human-translated into 31 languages. Because of the obscurity of the knowledge required to answer questions in \multiloko and \eclektic, even state-of-the-art LLMs have very low accuracy. From \gmmlu, we select the prehistory subset as those questions primarily involve direct knowledge-recall. \gmmlu is, however, a multiple-choice dataset, so we select only questions that suit free-form short response (see Appendix~\ref{gmmlu_details}).
Our final \gmmlu prehistory dataset consists of 150 questions in 42 languages each.

Since all three datasets constitute short-answer QA, we use \qwen as an autorater to evaluate the (binary) correctness of the model response (see Appendix~\ref{eval_details} for autorater details).

\section{Cross-Lingual Inconsistency} \label{inconsistency}

Our method depends on the prevalence of cross-lingual inconsistency. Such inconsistency
is often studied in the field of multilingual NLP as a problematic \textit{gap} between higher- and lower-resource languages, with many works leveraging interpretability methods \citep{lim2025languagespecificlatentprocesshinders,lu-etal-2025-paths,wang-etal-2025-lost-multilinguality, ai-etal-2025-knowledge}. Inconsistency is highest, of course, for challenging tasks. As LLMs have scaled and improved, however, this inconsistency has not been subdued, but rather shifted to new tasks \citep{qi-etal-2023-cross, hu-etal-2025-quantifying}. Inconsistency is more common in rarer information (i.e. data less prevalent in the training corpus) than facts like ``capital of France is Paris'' \citep{liu-etal-2025-tracing, romanou2025include}, where knowledge representations are less brittle.
Cross-lingual variance can also arise from simple run-to-run variance, which we find to be \textit{significantly} higher for lower-resource languages, in line with \citet{piratla2025rethinkingcrosslingualgapsstatistical}. 

When prompting an LLM with a factual question in many languages, three outcomes are possible:
\begin{enumerate}[leftmargin=*, nosep]
    \item \textit{Model is always wrong.} In such cases, attempting to localize the components that enable this knowledge is nonsensical.
    \item \textit{Model is always right.} Here, our method cannot be applied: a significant limitation. However, we argue that it requires impressive robustness for an LLM to consistently answer about a fact correctly across many languages. This robustness requires complex and redundant parameterization that would, from the start, make the task of knowledge localization very challenging in MoEs. As we show in Table~\ref{inconsistency_tbl}, this case almost never occurs on our datasets and models.
    \item \textit{Model is inconsistent.} This scenario provides the contrastive signal necessary for our statistical identification method.
\end{enumerate}

\noindent On these questions, we display the frequency of these three outcomes on our three datasets and two models in Table~\ref{inconsistency_tbl}. Note that given such difficult questions, it is very rare for these LLMs to be able to answer the question across all languages. In addition, many samples from these datasets we discard for our study because the LLMs display no evidence of holding that knowledge (All Incorrect).

\begin{table}[H]
    \centering
    \caption{Prevalence of cross-lingual inconsistency (``Mixed Results''). The first two columns give dataset statistics. Accuracy is averaged over all languages and questions. Statistically, (1) more languages and (2) accuracy closer to 50\% increase the chance of inconsistency.}
    \footnotesize
    \setlength{\tabcolsep}{2.5pt}
    \renewcommand{\arraystretch}{0.95}
    \begin{tabularx}{\linewidth}{>{\raggedright\arraybackslash}X | cc |ccccc|ccccc}
        \toprule
        & & & \multicolumn{5}{c|}{\textbf{\glm}} & \multicolumn{5}{c}{\textbf{\qwen}} \\
        \cmidrule(lr){4-8} \cmidrule(lr){9-13}
         & &  & \textbf{Eng} & \textbf{Avg}  & \textbf{All} & \textbf{All} & \textbf{Mixed} & \textbf{Eng} & \textbf{Avg}  & \textbf{All} & \textbf{All} & \textbf{Mixed} \\
        \textbf{Dataset} & \textbf{Q's} & \textbf{Langs} & \textbf{Acc.} & \textbf{Acc.} & \textbf{Corr.} & \textbf{Incorr.} & \textbf{Results} & \textbf{Acc.} & \textbf{Acc.} & \textbf{Corr.} & \textbf{Incorr.} & \textbf{Results} \\
        \midrule \midrule
        \eclektic & 383 & 12 & 28.5 & 21.5 & 9 & 172 & \textit{202} & 21.7 & 15.4 & 6 & 221 & \textit{156} \\
        \multiloko & 250 & 31 & 42.0 & 24.5 & 0 & 92 & \textit{158} & 33.6 & 15.8 & 0 & 111 & \textit{139} \\
        \gmmlu prehistory & 150 & 42 & 54.7 & 21.2 & 0 & 28 & \textit{122} & 61.3 & 16.3 & 0 & 26 & \textit{124} \\
        \bottomrule
    \end{tabularx}
    \label{inconsistency_tbl}
\end{table}

And so while the requirement of cross-lingual inconsistency limits the scope of facts XICI can be applied to, we argue that knowledge with such inconsistency is specifically the knowledge whose parameterization is brittle enough to be isolated to small number of components \citep{yu-etal-2023-characterizing}.
\section{Expert Identification Methodology} \label{identification}

XICI attributes knowledge to experts using a contrastive analysis of model routing when an LLM answers a question correctly versus incorrectly.

\subsection{MoE Preliminaries and Notation}

To selectively activate a subset of experts, a \textit{router}, or gating function, takes the inputs to an MoE block in a transformer decoder layer and produces logits (one for each available expert). The input hidden states are then sent only to the top-$k$ experts, whose outputs are typically aggregated using a normalized weighted mean.
We take notation from
\citet{bandarkar2026multilingual}:

\begin{itemize}[nosep]
    \item let $E$: the number of experts in each MoE layer.
    \item let $N$: the number of sequences in the corpus.
    \item let $L_i$: the length (number of tokens) of the $i^\text{th}$ sequence. A sequence consists of inputs and outputs.
    \item let $\bm{z}^{(\text{lang},l)}_{i,t} \in \mathbb{R}^E$: the router logits at the $t^\text{th}$ token of the $i^\text{th}$ sequence in language $\text{lang}$ at layer $l$.
\end{itemize}

\noindent Both \glm and \qwen have $E$$=$$128$ experts per layer and activate $k$$=$$8$ per token. \glm additionally has a \textit{shared} expert that is activated for all tokens and a singular dense (non-MoE) layer prior to 45 MoE decoder layers. Conversely, \qwen has all 48 MoE layers. For both models, the MoE blocks hold about 95\% of the total parameters.

\subsection{Routing Data Collection}

When evaluating the models on each question in each language, we track the router outputs at each MoE layer for each token in the full sequence (prompt, output, and even special tokens).
Rather than using the raw logits $\bm{z}$, we seek to determine the experts most \textit{responsible} for the outcome. As a result, we use each model's specific gating function $G$ to get: $\bm{p}^{(\text{lang},l)}_{i,t}$, the actual \textit{weight} of each expert's output when aggregating to compute the layer output. This is an $E$-dimensional probability (sum $=1$), with values at 0 for all experts not in the top-$k$. We discuss the models' gating functions and their consequences in Appendix~\ref{model_router_details}.

To reduce token-level noise, we average $\bm{p}^{(\text{lang},l)}_{i,t}$ across tokens $t$ in a sequence to get $\bm{p}^{(\text{lang},l)}_{i}$. To not make assumptions about \emph{when} the model fetches the parametric knowledge, we average over tokens in the entire sequence. This serves as our measure for how responsible each expert was for the model output. All in all:
\begin{equation} \label{gating}
\bm{p}^{(\text{lang},l)}_{i} = \frac{1}{L_i}\sum_{t=1}^{L_i}G(\bm{z}^{(\text{lang},l)}_{i,t}) \ \ \ \ \ \ \ \in [0,1]^E
\end{equation}


We implement this routing data collection based on \citet{fayyaz2026steering} (see Appendix~\ref{collection_appendix}). 


\subsection{Avoiding Generalist Experts} \label{generalist}

In order to localize knowledge based on the routing patterns, we first take three steps to avoid experts that are important for answering the style of question, but that are not question-specific. These could be experts important for general mechanisms, certain languages, low-level language syntax, or other.

\paragraph{A. Excluding Top and Bottom Layers} To isolate experts governing semantic knowledge rather than surface-level linguistic features, we exclude the initial and final layers. Bottom layers are primarily responsible for low-level language processing and top layers output formatting and token generation. This has been shown explicitly for MoEs \citep{muennighoff2025olmoe, bandarkar2026multilingual}.  In addition, abstract concepts and knowledge are not likely to be parameterized here \citep{meng2023massediting, skean2025layer,xiao-etal-2025-analyzing, li-etal-2025-decoding, yang2026ace}, so including them poses a high risk of selecting experts based on surface-level confounders. For \qwen, we exclude the first 6 and last 6 layers, while for \glm, we exclude the first and last 5 layers.
For GLM, the first layer is not MoE, so we exclude this one plus 5 more. See Appendix~\ref{layer_exclusion} for further discussion.
\paragraph{B. Blacklisting Top Experts} We create a ``blacklist'' of experts consisting of any one that, for more than 5\% of a dataset, is in the top 1\% of most highly weighed experts. Intuitively, if an expert is called upon so much for numerous questions, they likely hold a role that is knowledge-agnostic. Overall, this results in blacklisting about 5\% of all experts. Informal exploration suggested that stricter blacklist cutoffs reduced XICI's Ablation Success Rate (Table~\ref{results_tbl}, Row 7) more than the random question-shuffling rate (Row 11). Looser cutoffs increased both; consistent with the blacklist becoming less effective at excluding generalist experts, undermining the confidence in the localization method. We therefore chose a middle ground.
\paragraph{C. Subtracting Language-Specific Average} To further minimize confounding of language-specific parameterization in the selection of experts, we subtract the average weight across each language: 
\begin{equation}
    \Delta^{(\text{lang},l)}_{i} := \bm{p}^{(\text{lang},l)}_{i} - \frac{1} {N}\sum_{j=1}^N\bm{p}^{(\text{lang},l)}_{j}
\end{equation}

\subsection{Statistical MWU Test} \label{stattest}

The central and most discriminative step in XICI is a statistical test. For each question $i$, we partition languages into two groups based on model performance: correct ($C_i$) and wrong ($W_i$). For each expert $\varepsilon$ in each layer $l$, we seek to determine whether the distribution of values $\{\Delta^{(\text{lang},l)}_{i}(\varepsilon)~|~\text{lang} \in C_i\}$ is higher than $\{\Delta^{(\text{lang},l)}_{i}(\varepsilon)~|~\text{lang} \in W_i\}$. To do so, we apply a one-tailed Mann-Whitney U test (a.k.a. Wilcoxon rank sum test) with significance threshold $p \leq 0.05$. This nonparametric test is applicable irrespective of the group sizes $|C_i|$ or $|W_i|$, though a more even split leads to more statistical power.

\subsection{Magnitude Threshold} \label{thresh}

The above statistical test is rank-based, meaning that while it is robust, it does not consider the \textit{magnitude} of difference between the two sets. Furthermore, given that there are many thousand experts per model, a significance level of $0.05$ would lead to very high risk of false discoveries because of the multiple comparisons problem. As a result, we add the stipulation that the difference in \textit{median} between the two groups of $\Delta^{(\text{lang},l)}_{i}(\varepsilon)$ values is above a threshold $\tau$. We use median to lessen the sensitivity to outliers. Based on an analysis of routing data, we select $\tau = 0.01$ for \glm and $\tau = 0.005$ for \qwen (See Appendix~\ref{magthresh}). In a sense, these values, along with the "Top Experts" threshold in Section~\ref{generalist}, control the balance of false positives (identifying experts not specific to a question) to false negatives (not identifying experts important for the question).


\subsection{Expert Ranking} \label{ranking}

For a minority of questions, a high number of experts pass the 5 conditions outlined in Sections~\ref{generalist}--%
\ref{thresh}. We put a per-question cap of 25 on how many experts we select to further lower the risk of `false discoveries' and keep the results easy to interpret (See Appendix~\ref{maxexperts}). To determine these 25, we rank experts by both the U-statistic\footnote{Based on the \citet{bh} procedure for controlling false discovery with multiple hypothesis tests} and the magnitude of difference and select those with the lowest combined ranking. For most questions, this step is unnecessary (see Row 5 in Table~\ref{results_tbl}).


\section{Validating Attributed Experts} \label{intervention}

\subsection{Causal Ablation Methodology}

Verifying whether an LLM localization method reliably identifies model components is inherently difficult \citep{hase2023does,wei2024does,chang-etal-2024-localization}. The standard for verifying the functional necessity of localized components remains causal ablation. 
Fortunately, causally ablating experts in MoEs is relatively straightforward. 

We use the expert deactivation implementation from \citet{fayyaz2026steering}. During the MoE forward pass, it replaces the logit for the target expert $z(\varepsilon_t)$ with $\text{min}_\varepsilon z(\varepsilon)$. And because both models renormalize weights for the top-$k$ experts selected, this intervention ``mutes'' an expert while maintaining a natural distribution of output embeddings.\footnote{See Appendix~\ref{model_router_details} for discussion.}
Rather than ablating experts one by one, we perform a per-question intervention: all identified experts for a question are deactivated together. We then re-evaluate the model's response across all languages.

While similar causal ablation experiments would typically only evaluate on questions the model originally got right, we choose to evaluate on all questions and report three metrics. (1) Ablation Success Rate: For each question in each language, the percentage of question, language $(i, \text{lang})$ pairs where the model originally got right that it gets wrong after ablation. (2) Spurious Gain Rate: The percentage of $(i, \text{lang})$ pairs where the model originally got it wrong, and now gets it right. (3) \textcolor{orange!30!black}{Rate Difference}: This is simply the difference between the Ablation Success Rate and the Spurious Gain Rate. Note however, that given the low baseline accuracy of the dataset (e.g., 16\%), this is not equivalent to net accuracy drop. We argue this metric fairly penalizes methods that achieve high disruption through variance or general degradation.

\subsection{Random Baselines}
To contextualize ablation results on \methodname-identified experts, we calculate two random baselines:

\begin{itemize}[nosep, leftmargin=*]
\item \textbf{Random Expert Set of Same Size}
For each question evaluated, there are 1 to 25 experts that have been identified. For each question, we randomly sample experts to have a set of exactly the same size $\in [1,25]$. We do not sample from the layers originally excluded. If the Ablation Success Rate is high here, this implies the model simply breaks if this quantity of experts are deactivated at once. Note that, 25 experts $\approx0.3\%$ of all experts.

\item \textbf{Random Question-Shuffling}
To evaluate the specificity of our identification to individual pieces of knowledge, rather than knowledge in general, we deactivate the experts identified for question $r(i)$ while asking the model question $i$, where $r(i)$ is chosen by randomly pairing each $i$ with an $r(i) \neq i$ without repetitions, i.e. no two distinct $i$ map to the same $r(i)$. 
If the Ablation Success Rate is high after this shuffling, this implies these experts hold a broader role for answering that style of question rather than being fact-specific.
\end{itemize}

\subsection{Causal Ablation Results}

\begin{table}[t]
\centering
\small
\caption{Results from our causal ablation of attributed experts. Row 1 is the number of questions we apply \methodname~to, while Row 4 is the number of questions where experts were actually identified.}
\label{results_tbl}
\resizebox{\linewidth}{!}{%
\setlength{\tabcolsep}{2pt}
\begin{tabular}{@{}cl||ccc|c|ccc|c@{}} 
\toprule
& \textbf{Model} & \multicolumn{4}{c|}{\textbf{\glm}} & \multicolumn{4}{c}{\textbf{\qwen-Instruct-2507}} \\
& Details & \multicolumn{4}{c|}{45 MoE layers, 5760 total routed experts} & \multicolumn{4}{c}{48 MoE layers, 6144 total experts} \\
& Configuration & \multicolumn{4}{c|}{max 25, $\tau$ = 0.01, layers 6-40} & \multicolumn{4}{c}{max 25, $\tau$ = 0.005, layers 6-42} \\
\cmidrule(lr){3-6} \cmidrule(lr){7-10}
& \textbf{Dataset} & \textbf{\eclektic} & \textbf{\multiloko} & \textbf{G-MMLU} & \textbf{Total} & \textbf{\eclektic} & \textbf{\multiloko} & \textbf{G-MMLU} & \textbf{Total} \\ \midrule \midrule
\rownumber & Num. Q's w/ Inconsistency & 202 & 158 & 122 & 482 & 156 & 139 & 124 & 419 \\
\rownumber & Num. Languages & 12 & 31 & 42 & - & 12 & 31 & 42 & - \\
\rownumber & Blacklisted experts & 316 & 290 & 285 & - & 298 & 270 & 240 & - \\ \midrule
& \multicolumn{9}{l}{\textit{\methodname, our method}} \\
\rownumber & Num. Q's w/ Experts ID'd & 134 & 125 & 122 & 381 & 99 & 102 & 124 & 325 \\
\rownumber & Avg. Experts ID'd per Q & 21.9 & 15.2 & 24.0 & 20.4 & 16.9 & 9.1 & 22.8 & 16.7 \\
\rownumber & Orig. Correct \{lang, $i$\} & 715 & 1752 & 1322 & 3789 & 514 & 1005 & 1017 & 2536 \\ \midrule \midrule
\rownumber & Ablation Success Rate & 43.7\% & 50.1\% & 47.5\% & 48.0\% & 33.2\% & 34.5\% & 37.6\% & 35.5\% \\
\rownumber & Spurious Gain Rate & 5.3\% & 4.2\% & 3.4\% & 3.9\% & 5.0\% & 3.0\% & 3.4\% & 3.4\% \\
\rownumber & \textcolor{orange!30!black}{Rate Difference} & \textcolor{orange!30!black}{38.4\%} & \textcolor{orange!30!black}{45.9\%} & \textcolor{orange!30!black}{44.1\%} & \textcolor{orange!30!black}{44.1\%} & \textcolor{orange!30!black}{28.2\%} & \textcolor{orange!30!black}{31.6\%} & \textcolor{orange!30!black}{34.2\%} & \textcolor{orange!30!black}{32.0\%} \\
\rownumber & Num Q's All Incorrect & 24 & 28 & 17 & 69 & 14 & 10 & 14 & 38 \\
\midrule
& \multicolumn{9}{l}{\textit{Random Question-Shuffling (baseline)}} \\
\rownumber & Ablation Success Rate & 7.6\% & 6.6\% & 8.0\% & 7.3\% & 7.0\% & 7.3\% & 7.6\% & 7.4\% \\
\rownumber & Spurious Gain Rate & 4.6\% & 3.0\% & 4.0\% & 3.8\% & 4.4\% & 2.7\% & 4.2\% & 3.7\% \\
\rownumber & \textcolor{orange!30!black}{Rate Difference} & \textcolor{orange!30!black}{3.0\%} & \textcolor{orange!30!black}{3.6\%} & \textcolor{orange!30!black}{4.0\%} & \textcolor{orange!30!black}{3.5\%} & \textcolor{orange!30!black}{2.6\%} & \textcolor{orange!30!black}{4.6\%} & \textcolor{orange!30!black}{3.4\%} & \textcolor{orange!30!black}{3.6\%} \\
& \multicolumn{9}{l}{\textit{Random Expert Set of Same Size (baseline)}} \\
\rownumber & Ablation Success Rate & 7.5\% & 4.1\% & 4.8\% & 5.0\% & 6.1\% & 2.9\% & 5.1\% & 4.4\% \\
\rownumber & Spurious Gain Rate & 5.6\% & 3.4\% & 3.7\% & 3.9\% & 5.2\% & 1.8\% & 3.6\% & 3.2\% \\
\rownumber & \textcolor{orange!30!black}{Rate Difference} & \textcolor{orange!30!black}{1.9\%} & \textcolor{orange!30!black}{0.7\%} & \textcolor{orange!30!black}{1.0\%} & \textcolor{orange!30!black}{1.1\%} & \textcolor{orange!30!black}{0.9\%} & \textcolor{orange!30!black}{1.1\%} & \textcolor{orange!30!black}{1.5\%} & \textcolor{orange!30!black}{1.2\%} \\
\bottomrule
\end{tabular}%
}

\end{table}

\paragraph{Baselines} Rows 14-16 of Table~\ref{results_tbl} show that randomly deactivating this quantity of experts has non-negligible impact.
Crucially, however, our random question-shuffling baseline (rows 11-13) leads to only marginally more disruption.
We hypothesize that because multiple distinct pieces of knowledge could potentially be co-parameterized within the same expert, an Ablation Success Rate of only ~2-3\% above random selection indicates that \methodname is very rarely selecting "generalist" experts.

\paragraph{\methodname} In comparison to these randomized baselines, \methodname leads to a significantly higher success rate and rate difference (rows 7-9). Albeit far from 100\%, \methodname often identifies experts that are essential for answering that specific question. In general, the ablations reveal a higher success for \glm, where the rate difference is more than 40\% more than the random baselines, compared to about 30\% more for \qwen. This reflects the fact that we found fewer experts on average per question for Qwen3 (row 5). Row 10 shows the number of questions where, after ablation, the model can no longer answer it correctly in \textit{any} language. For \glm, we fully disable the model's knowledge in 69 out of 381 (18\%).

We also examined whether ablation effectiveness varied with the correct/wrong language split for each question. The relationship with Rate Difference was noisy, but ablation was markedly less effective when the model answered correctly in more than 80\% of languages (see Section~\ref{discussion}). One possible explanation is that knowledge expressed consistently across languages is more robust to ablation, though this remains speculative.


\subsection{Comparison to Using English Paraphrases} \label{engparaphrases}

Fundamentally, XICI leverages a model's inconsistent behavior across semantically equivalent inputs, but do these inputs really need to be in different languages? We evaluate the feasibility of instead using inputs in a single language by generating \textit{paraphrases} for each question in the English split of \multiloko using \gemini-3-Pro \citep{gemini3}. We carefully review the \gemini outputs and find that the largest number of consistently high-quality and semantically-equivalent paraphrases for the questions is approximately 10. With these 11 paraphrases (including the original) of each \multiloko question, we repeat the process outlined in Section~\ref{identification} to identify experts.

\begin{table}[H]
\centering
\caption{Comparison of using English paraphrases rather than Cross-lingual translations.}
\footnotesize
\setlength{\tabcolsep}{3pt}
\renewcommand{\arraystretch}{0.95}
\begin{tabularx}{\linewidth}{>{\raggedright\arraybackslash}X||c c|c c}
    \toprule
    Model & \multicolumn{2}{c}{\textbf{\glm}} & \multicolumn{2}{c}{\textbf{\qwen}} \\ \cline{2-5}
    Method (Num. Variants) & \textbf{Paraphrases (11)} & \textbf{Translations (31)} & \textbf{Paraphrases (11)} & \textbf{Translations (31)} \\
    \midrule \midrule
    All Correct & 70 & 0 & 37 & 0 \\ 
    All Incorrect & 104 & 92 & 88 & 111 \\ 
    Mixed Results & 76 & 158 & 125 & 139 \\
    \midrule
    Questions w/ Experts ID'd & 47 & 125 & 56 & 102 \\
    Avg. Experts ID’d per Q & 9.7 & 15.2 & 9.4 & 9.1 \\
    \midrule
    Ablation Success Rate & 34.5\% & 50.1\% & 18.2\% & 34.5\% \\ 
    Spurious Gain Rate & 2.3\% & 4.2\% & 2.3\% & 3.0\% \\ 
    Rate Difference & 32.2\% & 45.9\% & 15.9\% & 31.6\% \\ 
    \bottomrule
\end{tabularx}
\label{engpara_tbl}
\end{table}

As we show in Table~\ref{engpara_tbl}, using single-language paraphrases simply works less well, which we boil down to two main reasons. (1) The higher syntactic overlap between English paraphrases leads to less variation in performance. That is, the LLM is much more likely to always answer correctly or incorrectly across paraphrases. Additionally, internal expert activations are also less varied run-to-run. And (2) there are only so many same-language paraphrases possible (in comparison to languages you can translate to). As a result, fewer parallel questions means lower statistical power for XICI's MWU test. Accordingly, we can apply this method to fewer questions and are able to identify far fewer statistically significant experts. This implies that the experts important for factual recall are genuinely sensitive to surface input differences. However, in cases where you can identify a similar number of experts as with translations, we find that the ablation success rate is reasonable, albeit lower.
\section{Discussion} \label{discussion}

\paragraph{Contextualizing the Results} Our results can be summarized by a \textcolor{orange!30!black}{Rate Difference} (Ablation Success Rate $-$ Spurious Gain Rate) of 44.1\% and 32.0\% for the two models, respectively, across these datasets.
While expert-level knowledge localization has been explored in \citet{li-etal-2025-decoding}, a direct quantitative comparison is not possible as their method uses expert deactivation for localization itself. Meanwhile, \methodname~uses it solely for validating. Nonetheless, \methodname~specifically addresses (and is tested on) deep and sparse MoEs, which \citet{li-etal-2025-decoding} find are significantly more robust to expert deactivations. Given this, our ability to achieve such high Ablation Success Rate represents a promising step toward higher-fidelity knowledge localization in MoEs. At the same time, validating localization remains inherently difficult \citep{hase2023does,wei2024does,chang-etal-2024-localization}: causal ablation is only a proxy, and an ablation success rate of 40\% must be interpreted in light of the model, architecture, and facts involved. Moreover, our method identifies experts important for retrieval, but does not establish whether they literally ``store'' the fact or instead support some sub-computation in its retrieval. While the lack of a comparable baseline limits the contextualization of our results, it serves to emphasize XICI’s novelty in a relatively unexplored area of MoE interpretability.


\paragraph{Non-Negligible Spurious Gain Rates} As displayed in Table~\ref{results_tbl}, there is a consistent, non-negligible number of questions where the models did not get it right prior to our intervention but did after (“spurious gain”), even in the random baselines. This is consistently higher than the typical run-to-run variance that we find for each question (about 2\%, discussed in Appendix~\ref{eval_details}), and is highest in English. This tells us something beyond noise: when ablation disrupts a question's usual route, the model can sometimes still reach the correct answer via a different routing path than the one it originally used. If this kind of rerouting happens on questions where ablation flips wrong to right, there is no reason to think it would not also happen on questions where the model was already right. The difference is simply that we cannot observe it there, since the answer does not change either way. In other words, some of what we count as ablation "failures" (right before and after) may not indicate the ablated experts were unimportant, but rather that the model fell back on a redundant, untargeted pathway. This means our Ablation Success Rate could be largely underestimating XICI’s ability to identify important experts. Generally, this Spurious Gain Rate points to genuine expert redundancy that would make expert-level localization and validation via ablation intrinsically hard.

\paragraph{Easier Questions are Harder to Ablate} We initially hypothesized that a fact where there are 50\% of languages where it gets it right and 50\% wrong (i.e. maximum statistical power) would lead to the most ablation success. However, when the degree of inconsistency is plotted against the \textcolor{orange!30!black}{Rate Difference} metric, it is very noisy. The only convincing trend is a much lower Ablation Success Rate when there is a smaller number of wrong samples. To be concrete: when for a question, the LLM is correct in >80\% of languages, the ablation of identified experts is markedly less effective. We speculate this is because in those cases the knowledge is more robust and therefore harder to remove.

\paragraph{Knowledge Experts are Strongly Dispersed}
Across both models, the majority of experts identified for a specific question are unique to that question. In fact, we find that these knowledge experts are quite evenly spread across model layers. For \glm there is slightly higher prevalence in the late-middle layers (Layers 26-40), but no such trend exists for \qwen (See Appendix~\ref{locations}). In other words, the experts specifically specialized to answer a given factual question are spread out across the model layers. However, we note that our method specifically filters out ``generalist'' experts that are useful for many facts at once. Similarly, these filtered-out generalists are also spread out yet figure most in the late-middle layers of GLM. As a result, we conclude knowledge and related functionality is heavily dispersed across model layers.

\paragraph{Necessary, but not Sufficient} We briefly experiment with \textit{activating} identified experts for all language, as opposed to \textit{deactivating}. That is, increasing their router logits by a fixed amount to encourage the LLM to route to those experts more often, as in \citet{bandarkar2026multilingual}. However, these simple attempts did not lead to positive impact. We reason that nudging the router towards important experts likely does not provide the LLM with missing upstream features necessary for actually retrieving the knowledge in question \citep{geva-etal-2023-dissecting, yao2024knowledge}



\paragraph{Translation for more Statistical Power} As discussed in Section~\ref{engparaphrases}, the more languages we can evaluate, the more likely we find required cross-lingual inconsistency and the more statistical power we have for our identification method. And while in this work, we used off-the-shelf datasets that were already translated, future work could use machine translation to expand the identification effectiveness. Importantly, any arbitrary factoid can be framed as a short question-answer pair and be machine-translated into many languages. And the translations need not be perfect, either; syntactic errors that do not undermine the overall prompt would contribute to creating variation for analysis.




 
\section{Conclusion and Future Work}


In this paper, we introduce a novel knowledge localization framework that transforms a persistent challenge in multilingual NLP—cross-lingual inconsistency—into a diagnostic tool for sparse Mixture-of-Experts (MoE) architectures. 
Prior localization methods generally construct contrastive conditions through synthetic perturbations to prompts or internal activations. Instead, we exploit naturally occurring cross-lingual variation in model behavior to apply a contrastive method for analysis. We find that the syntactic and representational variance inherent in cross-lingual samples provides a strong signal for isolating knowledge parameterization.
Our results across \glm and \qwen demonstrate the efficacy of this approach; by deactivating a sparse subset of identified experts---representing approximately $0.3\%$ of the total model parameters---we achieved an ablation success rate of over $40\%$. Crucially, our careful methodological design ensures these identified experts are knowledge-specific rather than simply essential for broader functionality.

Overall, our experiments point to a number of hypotheses on knowledge in MoE LLMs. Facts seem to be encoded in small clusters of specialist experts, but those clusters seem to be backstopped by redundant alternative pathways. Additionally, the whole retrieval process depends on upstream representational state that such experts themselves do not control.

Ultimately, while the redundant and distributed nature of MoE architectures presents significant challenges for interpretability, our work suggests that expert-level abstraction provides a more digestible, verifiable, and scalable approach for understanding the internals of modern LLMs. Moving forward, we call for methods that can further increase signal-to-noise ratio in interpreting MoE routing, especially in reasoning-heavy models.
Like other localization methods that infer component importance from perturbations or internal activations, \methodname---and especially its ablation-based verification---applies only to relatively brittle (i.e., inconsistent) knowledge, and the field still lacks approaches that can localize and verify more robust knowledge.  Substantial progress on these fronts is needed to move from interpreting model knowledge to robustly controlling it.
In addition, we point to future work understanding the dynamics of knowledge acquisition during pretraining and the relationship with expert specialization.
Significant progress is still required to improve knowledge localization and eventually move beyond interpretation to robust control over model knowledge.

\bibliography{custom}
\bibliographystyle{iclr2026_conference}




\appendix

\vspace{30em}

\section{Conversion of \gmmlu prehistory subset} \label{gmmlu_details}


While there are 324 questions in the prehistory subject subset of \gmmlu, many are not suitable to be converted from multiple-choice questions to short-answer QA. We initially ask \gemini to label each question using the following prompt:

\begin{quote}
Attached is a TSV with questions and answers. What I want you to do is annotate whether or not the question is well-defined. That is, whether someone answering would realistically know that the answer is the expected answer. Please judge the question on its own, carefully. Could a human or LLM really provide the expected answer unambiguously ?
Important context is that the Q\&A are stripped from a multiple choice QA dataset. Many times, the answers are unanswerable or not well-defined unless the respondent has other options to eliminate. For example, this is unanswerable: ``Q: Artifacts \_\_\_\_\_\_\_\_\_; ecofacts \_\_\_\_\_\_\_\_\_\_.
A: are made by humans; exhibit traces of human activity.''
\end{quote}

This led to 185 questions that we then manually scan to assure \gemini's labels. We manually further eliminate numerous that are not well-defined and end with 150 questions.

\section{Evaluation Setup Details} \label{eval_details}

\eclektic and \multiloko are evaluated using an auto-rater setup, using vLLM \citep{vllm} for inference. We prompt the questions with the system instruction "In just a few words, answer in the language of the question." and each of the model's individuals chat formats. Then, the model output (and not its thinking trace, if it accidentally generated one) is initially checked for exact match with the provided golden answers. For \multiloko, in particular, there are numerous possible formattings of the answer provided, which we check. If exact match does not find the answer, the model output is provided to an auto-rater (implemented with \qwen) along with the golden answer. Our auto-rater system instruction requires a one-word answer ("YES" or "NO") and provides context about the question and answer.  Our auto-rater system instructions are thoroughly tuned via qualitative analysis (i.e. mostly manual checks) to ensure reliability across languages. We assume it is less reliable in low-resource languages, even if we see no evidence of it.

We estimate the run-to-run variance to be about 1-2\% when altering prompts slightly or simply generation randomness (our temperature is fixed to 0.1). The source of this variance is likely a combination of the model being evaluated and the autorater model.

All experiments are conducted on a single 8-GPU node.

\section{Details of Model Routers} \label{model_router_details}

The router of the MoE block produces $\bm{z} \in \mathbb{R}^E$. Then the input hidden state is sent to the $k$ experts with highest logit. For each expert $\varepsilon \in \text{top-}k$, let the output hidden state be $o_\varepsilon$. A renormalized set of probabilities based on the logits $\bm{p}^* \in [0,1]^k$ is used to form a weighted mean of the outputs. In Eq.~\ref{gating}, we refer to the gating function $G(\bm{z})$: 

\begin{equation}
G(\bm{z})_\varepsilon = \begin{cases}
\frac{f(\bm{z})_\varepsilon}{\sum{j \in \text{top-}k} \ f(\bm{z})_j} & \text{if } \varepsilon \in \text{top-}k \\
\ \ \ \ \ \ \ \ \ 0 & \text{otherwise}
\end{cases}
\end{equation}

Function $f(\cdot)$ is a monotonic function that depends on the model. The choice of $f(\cdot)$ has major implications on the impacts of our expert identification method from Section~\ref{identification}, but also very small implications for our causal intervention method from Section~\ref{intervention}.

\paragraph{\qwen} In many MoEs, notably Qwen, $f(\cdot)$ is the softmax operation: so $f(\bm{z}) = \text{softmax}(\bm{z})$. Then, as described in the gating function, the softmax-probabilities for the top-$k$ logits only are renormalized to sum to one and get $\bm{p}^*$. 

Since the logits of experts not activated influences the softmax function and, therefore, even the renormalized probabilities of the top-k. As a result, deactivating an expert by setting its logit to a very negative number (e.g., $-1 \times 10^6$) causes softmax to flatten the difference in weights between the top-$k$ experts. The expert deactivation implemented in \citet{fayyaz2026steering} and \citet{bandarkar2026multilingual}, which we adopt, rather sets the logit of an expert we want to deactivate to min$(\bm{z})$ prior to $f(\cdot)$. And so this does disrupt the ``natural'' weight ratio between the top-$k$, those works show it is subtle enough that it does not disrupt LLM functionality.

\paragraph{\glm} Meanwhile, \glm uses the sigmoid operator: $f(\bm{z}) = \sigma(\bm{z})$. This leaves generally much more evenly distributed weights in comparison to typically sharper softmax, which they argue improves performance \citep{glm}. And since it is applied individually to every logit, intervening and modifying a single logit (such as for deactivating) does not disrupt the end-ratios between the top-$k$ output hidden states $o_\varepsilon$. As a result, intervening into \glm guarantees fully ``natural'' distribution of output weights.

\section{Logit Collection Implementation} \label{collection_appendix}

Similar to \citet{fayyaz2026steering}, we implement logit collection by adding code to the vLLM \citep{vllm} source implementations of the MoE blocks to save the router logits to local files. However, our implementation achieves this using a ``forward hook'' class, attached to the MoE module using the PyTorch \verb|register_forward_hook()| function. This additionally requires disabling prefix caching and batched inference, which significantly slows down the forward pass. Even so, this vLLM implementation is still faster than doing batched inference with \verb|output_router_logits=True| in the native Transformers package.

\section{Layer Exclusion} \label{layer_exclusion}

Our intuition to exclude bottom and top layers for knowledge identification is explained in Section~\ref{generalist}. The quantity of layers to exclude, however, requires further analysis. For \qwen, we leveraged the visualizations in \citet{bandarkar2026multilingual} of the same model (notably Figures 1, 2, and 5). We replicate a similar multilingual MoE routing analysis for \glm and find the patterns to be similar. These graphs show that the first and last 5 layers for \glm and 6 for \qwen have significant language-specific routing, and so we exclude this amount. 

\section{Max Experts Identified} \label{maxexperts}

\begin{figure*}[ht] 
    \centering
    \includegraphics[width=\columnwidth]{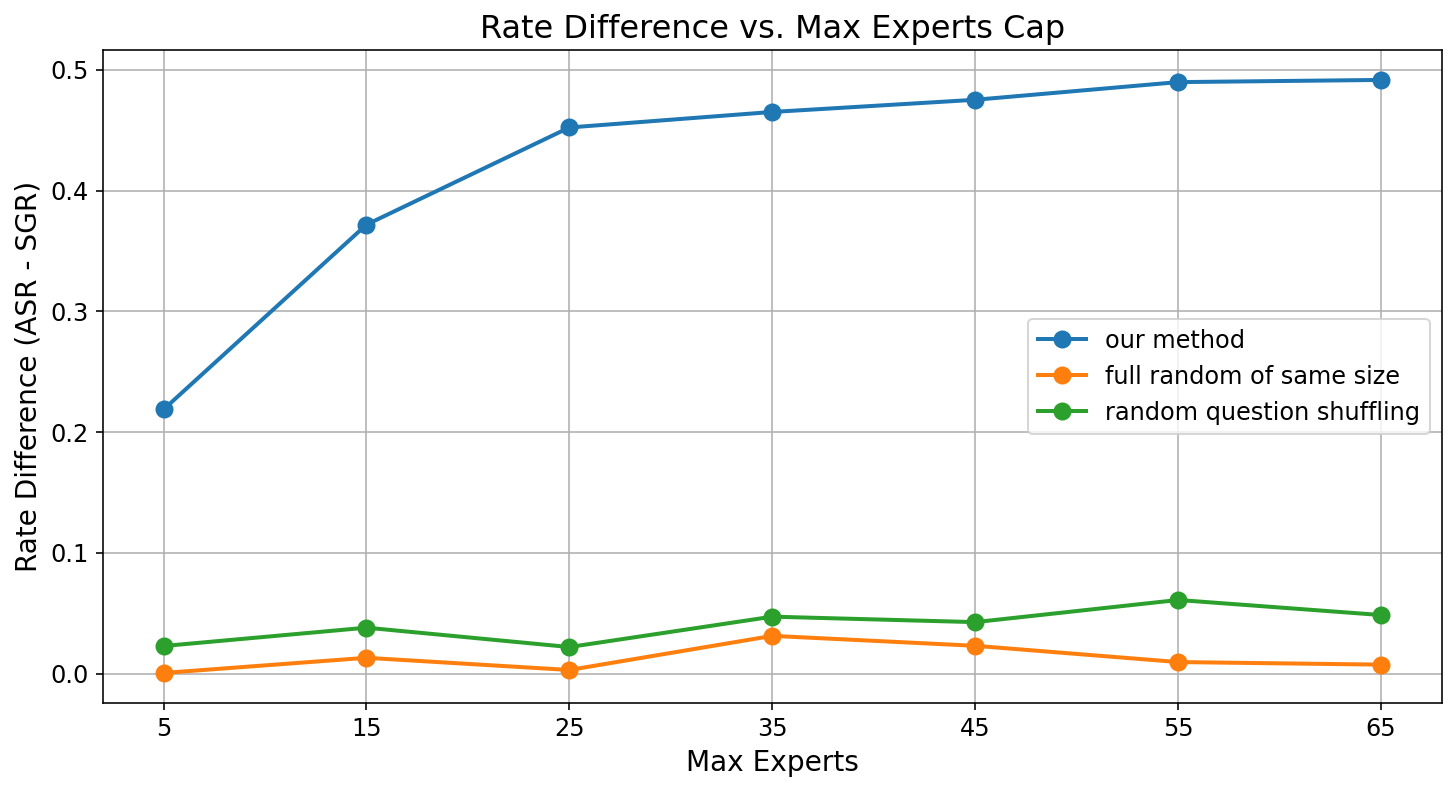}
    \caption{Impact of different ``max experts'' values for our method, when applied to \glm on \multiloko questions.}
    \label{fig:maxexperts}
\end{figure*}

As discussed in Section~\ref{ranking}, we set a maximum of 25 experts to be identified per question or piece of knowledge. This is a hyperparameter that can be tuned to moderate the false negative versus false positive ratio. As discussed, for most questions, less than 25 experts pass the statistical test so this maximum is only applied a minority of questions. In the chart in Figure~\ref{fig:maxexperts}, we illustrate what happens with different levels set (from 5 to 65).

We find that beyond 25 experts, the Ablation Success Rate, Spurious Gain Rate, and the same for the random baselines all flatten out. Partly, the difference in \textit{actual} experts selected is not big: even with a cap of 65, this cap is necessary only for 6.4\% of questions. Nonetheless, the small increase in average experts selected per question does not have a huge impact. This implies that these additional experts selected likely have little importance for answering the questions. Very curiously, the rate difference for the fully random baseline (of the same size) decreases after 35.

\section{Magnitude Threshold} \label{magthresh}

The threshold for determining a sufficient difference in \textit{median} delta values needed to be specific for each model, as the distribution of such values were quite different. This is likely related to the difference in gating functions, as discussed above in Appendix~\ref{model_router_details}. For \qwen and \glm, we tested out values 0.001, 0.003, 0.005, 0.01, 0.05. With higher values, too few experts were being selected (despite passing the U-test) and with lower values, too many were passing and the use of the inelegant max experts cap (as discussed in Appendix~\ref{maxexperts}) was called upon often.

\newpage

\section{Location of Identified Experts} \label{locations}

Below, we present a layer-wise distribution of the experts that we identified across all three factual-recall question-answering datasets using XICI. Note that the number of times identified is more than $E$$=$128 because an expert can be identified for more than one question (and each time it is identified counts towards the total).

\begin{figure*}[h]
    \centering
    \includegraphics[width=0.9\columnwidth]{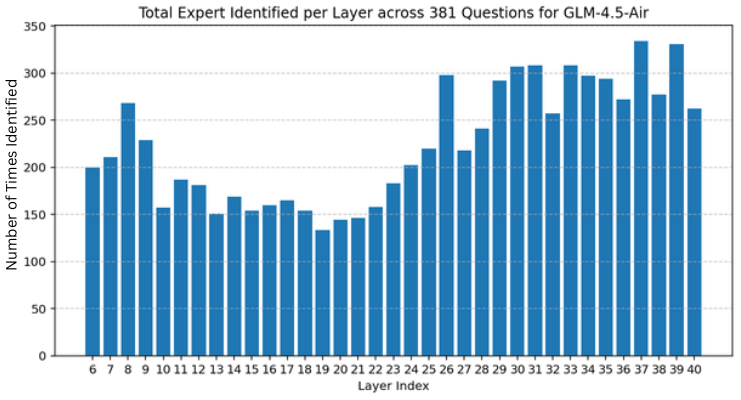}
    \caption{Location of experts identified for \glm.}
    \label{fig:glm_exp}
\end{figure*}

\begin{figure*}[h]
    \centering
    \includegraphics[width=0.9\columnwidth]{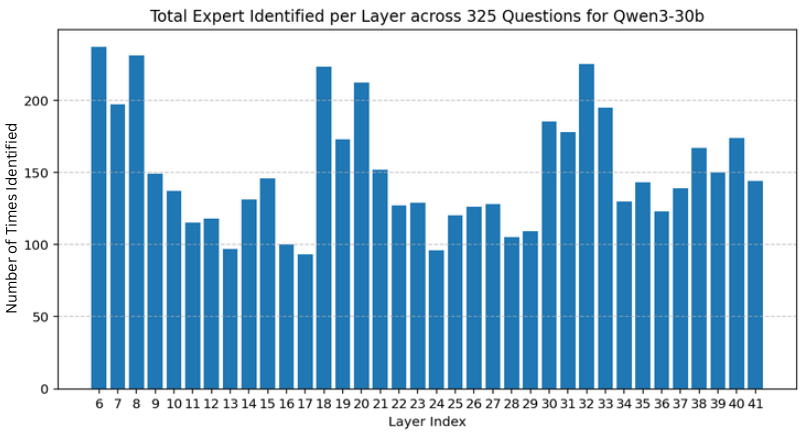}
    \caption{Location of experts identified for \qwen.}
    \label{fig:qwen_exp}
\end{figure*}

For \glm (Figure~\ref{fig:glm_exp}), we see a clear trend where experts in the back (a.k.a. top layers) are more often identified. However, this trend does not hold for \qwen (Figure~\ref{fig:qwen_exp}). For this model, rather, there are three distinct groups of layers where identification frequency spikes.

\section{Method Generalizability} \label{generalizability}

\paragraph{Thinking Models} This MoE routing analysis is applicable to thinking (or reasoning) LLMs, which, generate tokens for chain-of-thought processing prior to providing an answer. Thinking models perform significantly better on these datasets than their counterparts. However, we note that thinking tokens add a significant amount of routing noise, as the total sequence length increases dramatically. Verbose thinking typically hundreds of tokens and for this reason, we turn thinking off in this work.

\paragraph{Task Generalizability} 
While we experimented only with free-form, short-answer queries, the identification framework itself is fundamentally format-agnostic. We hypothesize that the method generalizes to any knowledge-dense task, though certain formats may introduce sub-optimal noise. For example, while the multiple-choice format (e.g., \gmmlu) could be adapted, we anticipate the presence of "distractor" options potentially dilutes the routing signal. Ultimately, this framework prioritizes tasks where the model's computations are dominated by the direct, parametric retrieval.

\section{Assets Licensing} \label{assets}

We use publicly released model checkpoints and benchmark datasets, adhering to their licenses:

\begin{itemize}[nosep, leftmargin=*]
    \item \qwen-Instruct-2507 \citep{qwen3}: Apache License 2.0.
    \item \glm \citep{glm}: MIT License.
    \item \multiloko \citep{hupkes2025multiloko}: MIT License. Note: The dataset repository additionally requires accepting its access conditions before downloading.
    \item \eclektic \citep{eclektic}: CC BY-NC-SA 4.0.
\end{itemize}



\end{document}